%% file: main_elsevier.tex
\documentclass[preprint,11pt,nopreprintline]{elsarticle}

\biboptions{numbers,sort&compress}
\journal{Image and Vision Computing}

\usepackage[left=25mm,right=25mm,top=25mm,bottom=25mm,paper=a4paper]{geometry}
\usepackage{amsmath,amsfonts}
\usepackage{algorithmic}
\usepackage{array}
\usepackage{textcomp}
\usepackage{url}
\usepackage{verbatim}
\usepackage{graphicx}
\usepackage{adjustbox}
\usepackage[permil]{overpic}
\usepackage{multirow}
\usepackage{booktabs}
\usepackage{xcolor}
\usepackage{subcaption}
\usepackage{lineno}
\usepackage{hyperref}

\newcommand{\method}{InfoTaxa}

\begin{document}

\begin{frontmatter}

\title{\method{}: Information-Calibrated Label-Free Clustering for \\ Fine-Grained Visual Taxonomy}

\author[csiro,qut]{David~Ahmedt-Aristizabal}
\author[csiro]{Mohammad~Ali~Armin}
\author[csiro,qut]{Abdelwahed~Khamis}
\author[qut]{Clinton~Fookes}
\author[qut]{Simon~Denman}
\author[csiro]{Lars~Petersson}

\address[csiro]{CSIRO, Australia}
\address[qut]{Queensland University of Technology, Australia}

\input{sec/0_abstract}


\end{frontmatter}

%
\input{sec/1_intro}

\input{sec/2_related}
\input{sec/3_method}
\input{sec/4_experiments}

\input{sec/5_discussion_limitations}

\input{sec/6_conclusion}


{
    \small
    \bibliographystyle{elsarticle-num}
    \bibliography{main}
}


\end{document}

%% file: sec/0_abstract.tex
\begin{abstract}
Label-free clustering of frozen visual representations can recover broad category structure, but poor fine-grained clustering does not reveal whether the limitation lies in the clustering algorithm or the representation itself. 
We present \textbf{\method{}}, \textit{an information-calibrated diagnostic} that combines image-only clustering performance, clustering efficiency relative to held-out probe-readable information, and predictive gain from a paired auxiliary modality. 
Clustering efficiency assesses the taxonomic information retained by the partition, while auxiliary gain measures additional predictive value beyond the tested image representation. 
On BIOSCAN-5M, a reference pipeline of BioCLIP~2, UMAP and HDBSCAN achieves adjusted mutual information (AMI) of $0.79$ at family but only $0.50$ at species. Its clustering efficiency decreases from $0.91$ at order to $0.53$ at species, while paired DNA reduces held-out species-level cross-entropy by $1.99$ bits. 
The DNA gain persists across encoder controls and barcode-grouped folds but decreases for better-sampled species. BioCLIP~2's pretraining exposure to the evaluation specimens limits generalisation claims; however, DINOv3, without explicit BIOSCAN taxonomic supervision, exhibits a similar pattern ($2.94$-bit DNA gain). 
On CUB-200-2011, image-derived attributes provide no measurable additional gain for six foundation-model encoders, but add 0.39 bits for ResNet-18.
Practical evaluations show that probe estimates support encoder comparison, but the diagnostic offers no clear advantage over simpler criteria across the three tested applications.
These findings indicate that species-level clustering on BIOSCAN-5M is both clustering-limited and representation-limited under the evaluated protocol, motivating improvements to clustering methods and visual representations.
\href{https://infotaxa.github.io/}{[\textbf{Project website}]}
\end{abstract}

%% file: sec/1_intro.tex
\section{Introduction}
\label{sec:intro}

\begin{figure*}
    \centering
    \includegraphics[width=1\linewidth]{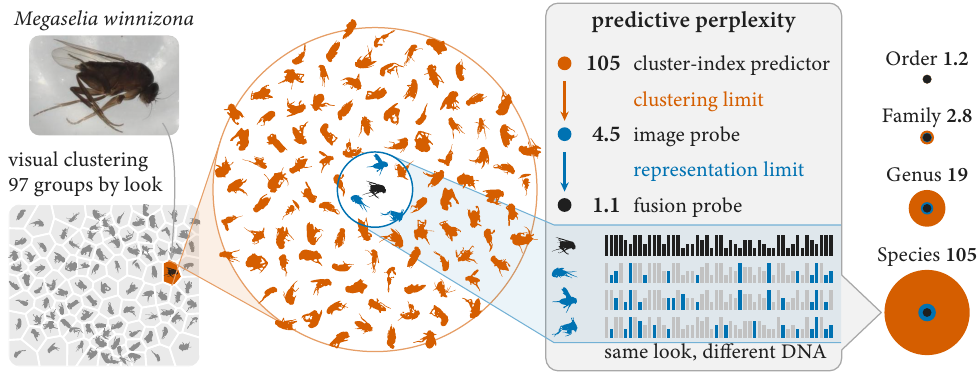}
    \caption{\textbf{Why does grouping insects by appearance fail at species level?}
    We find two reasons: a \emph{clustering limit} and a \emph{representation limit}.
    A BIOSCAN-5M specimen (\emph{Megaselia winnizona}) is placed in a visual cluster
    (BioCLIP~2 $\rightarrow$ UMAP $\rightarrow$ HDBSCAN).
    Each fly in the zoomed circle is a species it could be mistaken for. Numbers are species-level predictive perplexities, $2^{\mathrm{CE}}$ of the held-out cross-entropy (Table~\ref{tab:info}), for the cluster-index predictor, the image probe and the fusion probe.
    The image probe rules out most of them (orange):
    the information is there, but clustering discards it.
    The fusion probe, which adds DNA, rules out the rest (blue): look-alike species whose barcodes differ.
    Right: cluster-index perplexity is negligible at order and grows sharply towards species.
    }
    \label{fig:teaser}
\end{figure*}

Label-free clustering of frozen pretrained visual embeddings is an attractive option for scalable biodiversity monitoring. Images are inexpensive to collect, modern vision foundation models provide strong transferable representations, and cluster assignments can support taxonomic triage without exhaustive specimen-level annotation at inference~\cite{van2018inaturalist,oquab2024dinov2,stevens2024bioclip,gu2025bioclip2}. Yet image-only fine-grained taxonomy exhibits a consistent coarse-to-fine failure mode: visual embeddings recover broad taxonomic structure but struggle to extract correct grouping at finer levels, especially at the  species level~\cite{gharaee2024bioscan5m,lowe2024empirical,cole2022contrastive}.

This raises a central question: when fine-grained clustering fails, does the limitation arise from the clustering method or from taxonomic information that is not readily accessible in the frozen representation? If relevant structure is present but not recovered by the partition, alternative clustering or category-discovery methods may improve performance~\cite{van2020scan,vaze2022gcd,wen2023simgcd}. If the representation itself limits access to relevant taxonomic information, improving clustering alone may be insufficient. A clustering score by itself cannot distinguish these cases.


Fine-grained biological taxonomy provides a natural testbed because labels are hierarchical, from order and family to genus and species. BIOSCAN-5M is particularly useful in this respect because it provides paired specimen images, DNA barcodes, and Linnaean labels at scale~\cite{gharaee2024bioscan5m}. In this paper the inference pipeline remains image-only: paired DNA is used only post hoc, as an audit signal, to test where current visual embeddings stop exposing fine-rank taxonomic information (Figure~\ref{fig:teaser}).

To identify the limits of visual embeddings, we present \textbf{\method{}}, an information-calibrated diagnostic for label-free clustering of frozen visual representations. For each label level, \method{} reports three quantities: the adjusted mutual information (AMI) of an image-only partition; clustering efficiency, defined as the label information retained by the cluster index as a fraction of what a held-out probe reads from the same image embedding; and the held-out gain, $G^{\mathrm{fuse}}$, obtained by fusing a paired auxiliary signal with that embedding. Clustering efficiency measures clustering headroom, whereas the fusion gain audits the label signal beyond the frozen image representation. \method{} is not a new clustering algorithm or a new mutual-information estimator; rather, it combines these ingredients into a single diagnostic framework and tests whether the resulting diagnosis is actionable.

On the BIOSCAN-5M evaluation split, a simple reference image-only pipeline using BioCLIP~2 features~\cite{gu2025bioclip2}, UMAP reduction~\cite{mcinnes2018umap}, and HDBSCAN clustering~\cite{mcinnes2017hdbscan} reaches 0.79 AMI at family and 0.66 at genus, but only 0.50 at species. At order and family, the cluster index retains 0.91 and 0.82 of the label information that a held-out probe reads from the same embedding. At species, its efficiency drops to 0.53, and adding DNA to the image embedding yields a 1.99-bit held-out gain. These results indicate that, in the tested setting, species-level label-free clustering is both clustering-limited and representation-limited. Although BioCLIP~2's pretraining catalogue contains the evaluation specimens, a similar pattern is observed for DINOv3, which was not pretrained with explicit BIOSCAN taxonomic supervision. 
On CUB-200-2011, image-derived attributes provide no measurable additional gain for six foundation-model encoders, but improve held-out prediction for ResNet-18, demonstrating the dependence of the audit on the visual representation.
We further examine the utility of the diagnosis for encoder choice, clustering-method choice, and selective barcoding.
Our contributions are:
\begin{enumerate}
    \item \method{}, a diagnostic for label-free clustering of frozen visual representations that reports, for each label level, the partition score, clustering efficiency, and gain from a paired auxiliary signal.
    \item A full audit on BIOSCAN-5M, including controls for encoder choice, auxiliary choice, fold construction, class size, and label provenance, together with a contrasting audit on CUB-200-2011 using image-derived attributes.
    \item An evaluation of three potential uses of the diagnosis: encoder selection with limited labels, clustering-method selection, and selective DNA barcoding.
\end{enumerate}

%% file: sec/2_related.tex
\section{Related work}
\label{sec:related_work}

\noindent\textbf{Fine-grained visual recognition and biodiversity.}
Fine-grained recognition requires distinguishing visually similar categories, while biodiversity datasets introduce additional challenges through hierarchical taxonomy, long-tailed class distributions, and limited observations of rare species~\cite{van2018inaturalist,khan2023fishnet}. Although pretrained visual representations support recognition across biological categories, their ability to distinguish increasingly fine-grained classes remains limited~\cite{cole2022contrastive,gharaee2024bioscan5m}. Frozen representations have also been investigated for unsupervised analysis of biological imagery beyond insects~\cite{pastore2023efficient,ciranni2024plankton}, motivating methods that assess not only clustering performance but also the taxonomic information exposed by these representations.

\noindent\textbf{Label-free clustering with frozen visual features.}
Clustering methods differ in how they recover structure from visual representations. SCAN learns a clustering head following self-supervised pretraining~\cite{van2020scan}, while FINCH constructs partitions from nearest-neighbour relationships~\cite{sarfraz2019finch}. Lowe et al.~\cite{lowe2024empirical} systematically evaluate clustering on frozen pretrained representations, finding that UMAP reduction helps most clustering algorithms and that the choice among them has a small effect. Related evaluations on BIOSCAN-5M demonstrate the difficulty of recovering fine-grained taxonomic structure from image embeddings~\cite{gharaee2024bioscan5m}. These studies primarily characterise clustering performance, whereas \method{} examines how much of the information exposed by a frozen representation is retained by its partition and whether an auxiliary modality provides additional predictive information. Comparisons with supervised probes and representation-level diagnostics also motivate the evaluation of cluster quality relative to the information available in the underlying features~\cite{gadetsky2024turtle,adaloglou2023temi,laina2022measuring}.

\noindent\textbf{Category discovery and generalised discovery.}
Generalised category discovery (GCD) extends clustering by leveraging labelled examples to discover previously unseen categories~\cite{vaze2022gcd,wen2023simgcd,he2025categorydiscovery}. These approaches may use known or estimated category counts~\cite{vaze2022gcd,zhao2023gpc}, making their supervision and evaluation settings distinct from label-free clustering. We therefore distinguish methods that discover the number of clusters from those receiving oracle class counts or labelled examples, and include SimGCD as a semi-supervised reference rather than a directly comparable label-free baseline.

\noindent\textbf{DNA barcoding as an auxiliary audit signal.}
DNA barcoding supports species identification when morphological differences are subtle or cryptic~\cite{hebert2003barcodes,bickford2007cryptic}. Barcode index numbers (BINs) provide operational groupings derived from sequences~\cite{ratnasingham2013bin}, although barcode-based assignments are not error-free and should not be treated as independent taxonomic ground truth~\cite{meier2006diptera}. Learned genomic representations, including HyenaDNA~\cite{nguyen2024hyenadna}, enable DNA-based prediction, while CLIBD aligns image, DNA, and text embeddings~\cite{gong2024clibd}. 
DNA has also been used as additional semantic information for recognising unseen insect species~\cite{badirli2021dna,badirli2023unknown}. 
Unlike approaches that integrate DNA into recognition, \method{} uses paired DNA post hoc to audit predictive information beyond frozen image embeddings, without using it to construct image-only clusters. This distinction is relevant to biodiversity workflows in which image-based assessment may guide subsequent molecular identification~\cite{wuhrl2022diversityscanner}.

\noindent\textbf{Information-theoretic diagnostics and representation evaluation.}
Information-theoretic probing characterises the predictive information exposed by learned representations~\cite{pimentel2020information,voita2020information}. Usable, or $\mathcal{V}$-, information formalises this notion relative to a restricted predictor family~\cite{xu2020usable}, while conditional probing evaluates information available beyond a baseline representation~\cite{hewitt2021conditional}. These estimates depend on probe capacity and evaluation protocol, and should not be interpreted as exact mutual information~\cite{mcallester2020formallimitations}. Similarly, cluster--label information can be biased when estimated from partitions with many small groups~\cite{vinh2010information,romano2016adjusting}. Information-bottleneck formulations have considered the fraction of label-relevant information retained by a partition~\cite{slonim1999agglomerative,tishby1999information}, while related principles have informed fine-grained representation learning~\cite{ye2024r2trans}. Separately, transferability estimation methods assess the suitability of pretrained features for downstream recognition~\cite{you2021logme,pandy2022gbc,agostinelli2022stable}. Building on these established ideas, \method{} combines held-out probe estimates, clustering efficiency, and auxiliary predictive gain to diagnose fine-grained clustering limitations. We additionally examine the use of probe estimates for encoder selection, without claiming a new transferability score or superiority over methods not directly compared.

%% file: sec/3_method.tex
\section{The \method{} Diagnostic}
\label{sec:method}


\begin{figure*}[!t]
    \centering
    \includegraphics[width=0.99\linewidth]{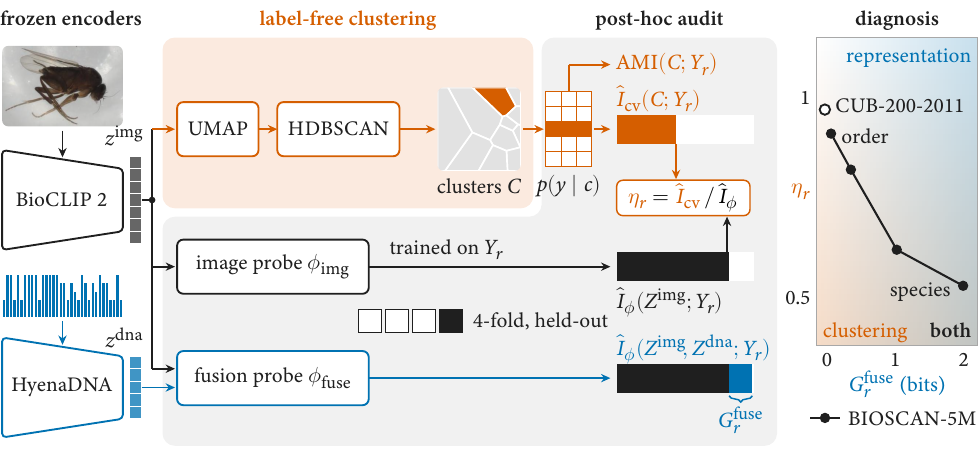}
\caption{\textbf{The InfoTaxa diagnostic.}
Frozen encoders embed each image and its paired DNA barcode. Label-free clustering (orange) uses image embeddings alone and is scored with AMI. The audit compares held-out label information from the cluster index, image probe $\phi_{\mathrm{img}}$, and fusion probe $\phi_{\mathrm{fuse}}$. Clustering efficiency $\eta_r$ indicates potential clustering headroom, while DNA fusion gain $G_r^{\mathrm{fuse}}$ indicates additional predictive value beyond the image embedding. Right: BioCLIP~2 results across BIOSCAN-5M taxonomic ranks and at species level on CUB-200-2011 (Tables~\ref{tab:info} and~\ref{tab:cub}; BIN overlaps species). Shading is qualitative.
}
    \label{fig:method_overview}
\end{figure*}

\method{} diagnoses the limitations of label-free clustering using frozen image embeddings at a given level of granularity. It takes a frozen image encoder, a dataset with labels at one or more granularities, and, where available, a paired auxiliary signal corresponding to the same items. For each label level, it reports the image-only partition score, clustering efficiency relative to held-out probe-readable information, and the held-out gain obtained by adding the auxiliary signal to the image embedding. Labels and the auxiliary signal are used for these post-hoc measurements, while the reference partition is formed from image embeddings alone. Figure~\ref{fig:method_overview} summarises the framework.

Although we primarily consider BIOSCAN-5M in this paper, the framework is more general. It applies whenever three ingredients are available: a frozen representation whose partition is being diagnosed, categorical or hierarchical labels for evaluation, and an auxiliary paired modality that can be used post hoc as an audit signal. In this paper, the auxiliary signal is DNA barcodes on BIOSCAN-5M, and annotated visual attributes on CUB-200-2011.

\subsection{Problem setup}
\label{sec:method_setup}
We consider a corpus
\begin{equation}
\mathcal{D}=\{(x_i,s_i,y_i^1,\ldots,y_i^R)\}_{i=1}^{N},
\end{equation}
where $x_i$ is an image, $s_i$ is a paired auxiliary record, and $y_i^r$ is the label at level $r$. On BIOSCAN-5M, $s_i$ is a DNA barcode and the levels are order, family, genus, species, and barcode index number (BIN). On CUB-200-2011, $s_i$ is a vector of annotated visual attributes. Because BINs are barcode-derived, species is the main fine-grained biological endpoint on BIOSCAN-5M and BIN is treated as an additional diagnostic.

\subsection{Image-only clustering pipeline}
\label{sec:method_clustering}
For each image encoder, we extract and cache frozen visual embeddings
\begin{equation}
z_i^{\mathrm{img}} = f_{\mathrm{img}}(x_i).
\end{equation}
No encoder is fine-tuned on the target dataset. The image-only inference path then reduces each embedding and assigns a cluster label,
\begin{equation}
c_i = h\!\left(g\!\left(z_i^{\mathrm{img}}\right)\right),
\end{equation}
where $g(\cdot)$ is a dimensionality-reduction operator and $h(\cdot)$ is an unsupervised clustering algorithm.

We use \textit{label-free clustering} to mean that neither target labels nor the paired auxiliary record is used to fit $g(\cdot)$ or $h(\cdot)$ for a given configuration. This does not imply unsupervised encoder pretraining or label-independent hyperparameter selection. Our reference pipeline uses UMAP~\cite{mcinnes2018umap} followed by HDBSCAN~\cite{mcinnes2017hdbscan}, which discovers the number of clusters from the embedding geometry. Alternative clustering methods are compared in Sec.~\ref{sec:exp_image_only}.

\subsection{Measured quantities}
\label{sec:method_probe}

\noindent\textbf{Partition score.}
For each label level $r$, we measure adjusted mutual information (AMI)~\cite{vinh2010information} between the cluster assignments $C$ and the labels $Y_r$.

\smallskip
\noindent\textbf{Probe-readable information.}
For a modality $m \in \{\mathrm{img},\mathrm{aux}\}$, we train a probe $\phi$ (denoted $\phi_{\mathrm{img}}$ for the image embedding) from a fixed predictor family $\mathcal{V}$ to predict $Y_r$ from $Z^m$, and evaluate it on held-out folds:
\begin{equation}
\widehat{I}_{\phi}(Z^m;Y_r)
=
\widehat{H}(Y_r)
-
\mathrm{CE}_{\phi}(Y_r \mid Z^m),
\end{equation}
where $\widehat{H}(Y_r)$ is the label entropy in bits, estimated with a Miller--Madow correction~\cite{miller1955note}, and $\mathrm{CE}_{\phi}$ is the mean held-out cross-entropy in bits. Following Xu et al.~\cite{xu2020usable}, we interpret this as the label information that the chosen probe family can read from the representation. In the main experiments, the probe is a shallow MLP; wider, linear, and non-parametric probes are used as diagnostics. The auxiliary embedding is
\begin{equation}
z_i^{\mathrm{aux}} = f_{\mathrm{aux}}(s_i).
\end{equation}
On BIOSCAN-5M, $f_{\mathrm{aux}}$ is a DNA encoder; on CUB-200-2011, it is the identity map on the attribute vector.

\smallskip
\noindent\textbf{Auxiliary gain.}
To test whether the auxiliary signal contains label information beyond the image embedding, we train a late-fusion probe, $\phi_{\mathrm{fuse}}$, that receives both modalities and measure
\begin{equation}
G_r^{\mathrm{fuse}} = \operatorname{CE}_{\phi_{\mathrm{img}}}(Y_r \mid Z^{\mathrm{img}})
-
\operatorname{CE}_{\phi_{\mathrm{fuse}}}\!\left(Y_r \mid [Z^{\mathrm{img}},Z^{\mathrm{aux}}]\right),
\end{equation}
with both terms evaluated out of sample. 
A positive $G_r^{\mathrm{fuse}}$ means that the auxiliary signal reduces held-out cross-entropy relative to image alone. We interpret this as a held-out audit of the label signal beyond the frozen image embedding, not as an exact conditional mutual-information estimate.

For a supporting comparison, we also report the marginal auxiliary--image probe difference
\begin{equation}
\Delta_r
=
\widehat{I}_{\phi}(Z^{\mathrm{aux}};Y_r)
-
\widehat{I}_{\phi}(Z^{\mathrm{img}};Y_r).
\end{equation}
Because $\Delta_r$ is a difference of two marginal probe estimates, it is not a conditional quantity and is reported only as supporting evidence.

\subsection{Clustering efficiency}
\label{sec:method_eta}
To compare what a partition captures with what a supervised probe can read from the same image embedding, we define clustering efficiency as
\begin{equation}
\eta_r =
\frac{\widehat{I}_{\mathrm{cv}}(C;Y_r)}
{\widehat{I}_{\phi}(Z^{\mathrm{img}};Y_r)}.
\end{equation}
The denominator is the held-out probe-readable information in the image embedding. The numerator, $\widehat{I}_{\mathrm{cv}}(C;Y_r)$, is the label entropy minus the held-out cross-entropy of a predictor that uses only the cluster assignment. For each training fold, label probabilities are estimated from cluster--label frequencies using one pseudocount per cluster, distributed uniformly across the $K_r$ classes:
\begin{equation}
p(y\mid c)=\frac{n_{cy}+1/K_r}{n_c+1},
\end{equation}
where $n_{cy}$ and $n_c$ denote the training-fold counts for label $y$ in cluster $c$ and the total count of cluster $c$, respectively. Both information estimates are evaluated on the same held-out folds (three four-fold splits) and expressed in bits.

High $\eta_r$ indicates that the partition retains taxonomic association comparable to that accessible to the image probe, while low $\eta_r$ indicates unrecovered probe-readable structure. This ratio is an empirical diagnostic, not a formal information bound, and is not necessarily restricted to $[0,1]$. The numerator and denominator use different predictors, a smoothed lookup table over clusters and a parametric probe, so $\eta_r$ can exceed one. Moreover, UMAP and HDBSCAN are fitted transductively on the full evaluation set, whereas probes are fitted on training folds only. The estimates therefore depend on the predictors, smoothing, and evaluation protocol. Provided that the image-probe information is positive and sufficiently far from zero, we use $\eta_r$ to assess potential clustering headroom across ranks, encoders, and datasets under a common protocol.

For intuition, a held-out cross-entropy CE can be read as a predictive perplexity, $2^{\mathrm{CE}}$, the effective number of equally likely candidate classes. Figure~\ref{fig:teaser} reports this species-level predictive perplexity for BioCLIP~2: about 105 for the cluster-index predictor of the reference partition, 4.5 for the image probe and 1.1 for the fusion probe (Table~\ref{tab:info}).

\begin{table}[t]
\centering
\caption{Qualitative interpretation of the \method{} diagnostic at label level $r$. The combinations suggest potential limitations and interventions rather than definitive decision rules.}
\label{tab:reading}
\resizebox{0.8\linewidth}{!}{%
\begin{tabular}{lll}
\toprule
 & Small or non-positive $G^{\mathrm{fuse}}_r$ & Large positive $G^{\mathrm{fuse}}_r$ \\
\midrule
High $\eta_r$ & Limited measured headroom & Potential representation limitation: \\
              & and auxiliary gain & assess encoder or auxiliary modality \\
\addlinespace
Low $\eta_r$  & Potential clustering limitation: & Both limitations indicated: \\
              & assess alternative partitions & assess clustering and representation \\
\bottomrule
\end{tabular}%
}
\end{table}

\subsection{Interpreting the diagnosis}
\label{sec:method_reading}
The two derived quantities provide complementary diagnostics: $\eta_r$ assesses how much probe-readable structure is retained by the partition, while $G_r^{\mathrm{fuse}}$ measures the additional predictive value of the paired auxiliary signal. Table~\ref{tab:reading} summarises their qualitative interpretation. Low $\eta_r$ suggests potential clustering headroom, whereas a large positive $G_r^{\mathrm{fuse}}$ motivates investigating alternative representations or additional measurements. When both conditions occur, as observed at species level on BIOSCAN-5M, the results indicate both clustering and representation limitations. Conversely, high $\eta_r$ and a small or non-positive $G_r^{\mathrm{fuse}}$ suggest limited headroom relative to the tested probe and auxiliary signal. These interpretations depend on the evaluation protocol and do not prescribe an intervention guaranteed to improve performance.

\subsection{Validity controls}
\label{sec:method_robustness}
Because the diagnosis depends on the probe and on the construction of held-out folds, we evaluate several controls. On BIOSCAN-5M, these include: a label-permutation baseline; a second auxiliary encoder; an auxiliary signal permuted across items or replaced by noise; folds that keep items with identical auxiliary records together; restrictions by class size and label provenance; alternative probe families and capacities; and multiple image encoders, including encoders with different forms of biological supervision. On CUB-200-2011, we use the corresponding controls that apply to the attribute-based auxiliary signal. Together, these controls test whether the measured auxiliary gain and clustering efficiency are robust to estimator choice, label construction, and dataset structure.

%% file: sec/4_experiments.tex
\section{Experiments}
\label{sec:experiments}

\subsection{Experimental setup}
\label{sec:exp_setup}

\noindent\textbf{Datasets and taxonomic labels.}
Our primary evaluation uses the cropped 256 split of BIOSCAN-5M~\cite{gharaee2024bioscan5m}, comprising 39,373 test and 7,887 test-unseen images (47,260 specimens in total). Each specimen has an image, a DNA barcode, and taxonomic labels from order to species, with BIN assignments available for all but 36 specimens. The combined split contains 27 orders, 355 families, 1,819 genera, 4,363 species, and 5,296 BINs. At species level, class sizes range from 3 to 33 specimens, with a median of 7 and a mean of approximately 11 specimens per species; 60\% of species have fewer than ten specimens. Species is our main fine-grained biological endpoint, while barcode-derived BINs provide an additional diagnostic.

BIOSCAN-5M includes both established species names and operational placeholder labels. Close to the dataset's curation convention, we identify placeholders using label-format criteria, keeping as described species the names of two or more words with no open-nomenclature epithet, numeral or BOLD identifier~\cite{gharaee2024bioscan5m}. These criteria support the described-species control but do not independently verify taxonomic provenance. Two additional properties are relevant to the DNA audit: 73.6\% of evaluation specimens share an identical barcode with another specimen, and some labels were harmonised across specimens sharing the same barcode. We examine the effects of barcode duplication and label provenance in Sec.~\ref{sec:exp_robustness}.

\noindent\textbf{Additional evaluation settings.}
We use three additional settings to examine the scope and stability of the diagnostic. The key-unseen partition of BIOSCAN-5M contains 36,465 specimens disjoint from the evaluation split, although many species overlap; it therefore tests transfer to different specimens rather than entirely unseen species. 
CUB-200-2011~\cite{wah2011caltech} contains 11,788 images of 200 bird species and 312 binary visual attributes annotated from the images, with known annotation noise~\cite{koh2020concept}. We use the combined official training and test partitions for four-fold cross-validation.
At species level, class sizes range from 41 to 60 images, with a median of 60 and a mean of approximately 59 images per species.
We obtain higher-level taxonomy from TreeOfLife-200M~\cite{gu2025bioclip2} and use the attributes as a paired auxiliary signal. FishNet~\cite{khan2023fishnet} and NEON carabid beetle data~\cite{east2026sentinelbeetles} provide additional image-only evaluations, but lack a paired auxiliary record for the full audit.

\noindent\textbf{Visual and auxiliary representations.}
All encoders remain frozen. BioCLIP~2 ViT-L/14~\cite{gu2025bioclip2} is our primary biological image encoder, while DINOv2 ViT-L/14~\cite{oquab2024dinov2} provides a non-biological visual control. We additionally evaluate CLIBD~\cite{gong2024clibd}, DINOv3~\cite{simeoni2025dinov3}, SigLIP~\cite{zhai2023siglip}, MetaCLIP~\cite{xu2024metaclip}, and EVA02~\cite{fang2024eva02}. DINOv2, DINOv3, and MetaCLIP use class-token features, SigLIP uses attention pooling, and EVA02 uses mean-pooled patch tokens. 
On CUB-200-2011, we additionally evaluate ImageNet-1k-pretrained ResNet-18 using 512-dimensional pooled features.
DNA sequences are encoded using HyenaDNA~\cite{nguyen2024hyenadna}, with the CLIBD DNA tower providing an alternative representation.

Pretraining exposure differs across encoders. BioCLIP~2 was trained with taxonomic supervision on TreeOfLife-200M, which includes BIOSCAN-5M. Matching specimen identifiers against the training catalogue identifies all 47,260 evaluation specimens, so its results should not be interpreted as generalisation to specimens unseen during pretraining. The catalogue also contains images belonging to all 200 CUB species, although this does not establish that the CUB evaluation images themselves were included. CLIBD was pretrained with paired image--DNA data from BIOSCAN-5M. The remaining encoders were not trained with explicit taxonomy from these corpora, although possible image overlap with large-scale web pretraining data cannot be excluded.

\noindent\textbf{Clustering configurations.}
Our reference image-only pipeline, denoted protocol \textsc{H}, applies UMAP-50~\cite{mcinnes2018umap} to BioCLIP~2 embeddings followed by HDBSCAN~\cite{mcinnes2017hdbscan,campello2013density}, with minimum cluster size 100 and minimum samples 5. UMAP uses 30 neighbours and zero minimum distance. 
These settings were inherited from earlier analyses comparing minimum cluster sizes of 30, 100 and 300, using family-level AMI on non-noise specimens and considering clustering coverage.
Thus, label-free refers to fitting the image-only partition without labels or DNA, rather than to an entirely label-independent selection of hyperparameters. 
On CUB-200-2011, the minimum cluster size is reduced to 25, preserving approximately the same fraction of the dataset and fixed before the evaluation reported here; an earlier exploratory run had compared sizes 10, 30 and 100 with labels in view. 
This proportional scaling does not account for differences in the ground-truth class-size distributions. 
In particular, the minimum cluster size of 100 on BIOSCAN-5M exceeds the largest species class (33 specimens) and the size of 81\% of families, constraining the granularity of the reference partition.
The alternative clustering methods evaluated in Sec.~\ref{sec:exp_image_only} provide complementary assessments without this minimum cluster-size constraint.
For comparison, Leiden~\cite{traag2019leiden} uses resolution 1 (the default, not tuned) on a nearest-neighbour graph of the reduced embeddings (29 neighbours per specimen); k-means and Ward are evaluated on the same representations.



\noindent\textbf{Evaluation protocol and baselines.}
We report adjusted mutual information (AMI)~\cite{vinh2010information} with arithmetic normalisation. All clustering scores include every specimen, treating HDBSCAN noise assignments as an additional cluster unless stated otherwise; we denote this metric as AMI(global). We distinguish this from the true-label entropy normalisation used for rank-wise results in prior BIOSCAN evaluations~\cite{gharaee2024bioscan5m,lowe2024empirical}, and compare the protocols explicitly in Sec.~\ref{sec:exp_image_only}.

Probe estimates use a shallow MLP with one hidden layer of 1,024 units per input block, batch normalisation, ReLU, dropout of 0.1, and a linear classification head. Features are L2-normalised and standardised using training-fold statistics. We train with AdamW (learning rate $2\times10^{-3}$, weight decay $10^{-4}$), cosine scheduling, and batch size 4,096 for 80 epochs under four-fold cross-validation. The primary evaluation uses random folds, with barcode-grouped folds as a robustness control. Probe results are averaged over three seeds and clustering results over five UMAP seeds; reported standard deviations reflect seed variability rather than uncertainty over independently sampled datasets.

Clustering comparisons include methods that discover $K$ (HDBSCAN and Leiden); methods using oracle class counts (k-means, Ward, and FINCH ~\cite{sarfraz2019finch}); and learned clustering heads, also given each rank's class count (SCAN~\cite{van2020scan}, TURTLE~\cite{gadetsky2024turtle}, and TEMI~\cite{adaloglou2023temi}). SimGCD~\cite{wen2023simgcd}, trained with all labels of a random half of the classes at each rank, serves as a semi-supervised reference rather than a directly comparable label-free baseline.

\begin{table}[t]
\centering
\caption{Image-only clustering on the BIOSCAN-5M evaluation split, measured using AMI(global). Numbers in parentheses in the column headings indicate the ground-truth class counts at each taxonomic level. The first two rows follow the published Ward protocol~\cite{gharaee2024bioscan5m}, using $K=4{,}363$ at every rank, on L2-normalised features. Oracle-$K$ methods and learned clustering heads use rank-specific class counts, while FINCH selects the nearest hierarchy level. Protocol \textsc{H} and Leiden do not require the class count; SimGCD is trained with the labels of a random half of the classes at each rank and serves as a semi-supervised reference. The fixed-$K$ rows and DINO oracle-$K$ row are averaged over three UMAP seeds, and other seeded rows over five (standard deviations $\leq$ 0.012). Results for learned clustering heads and SimGCD are obtained from single runs.}
\label{tab:methods}
\resizebox{0.8\linewidth}{!}{%
\begin{tabular}{llccccc}
\toprule
Features & Partition
& \shortstack{Order\\(27)}
& \shortstack{Family\\(355)}
& \shortstack{Genus\\(1,819)}
& \shortstack{Species\\(4,363)}
& \shortstack{BIN\\(5,296)} \\
\midrule
DINO ViT-B/16 & Ward, $K{=}4{,}363$ & .26 & .36 & .29 & .22 & .21 \\
BioCLIP~2 & Ward, $K{=}4{,}363$ & .29 & .52 & .58 & .53 & .52 \\
\midrule
DINO ViT-B/16 & Ward, oracle $K$ & .50 & .51 & .34 & .22 & .21 \\
BioCLIP~2 & Ward, oracle $K$ & .59 & .73 & .65 & .53 & .51 \\
BioCLIP~2 & k-means, oracle $K$ & .59 & .73 & .63 & .48 & .45 \\
BioCLIP~2 & FINCH, oracle level & .60 & .74 & .67 & .52 & .51 \\
BioCLIP~2 & TEMI~\cite{adaloglou2023temi} & .34 & .72 & .68 & .55 & .52 \\
BioCLIP~2 & SCAN~\cite{van2020scan} & .53 & .70 & .63 & .53 & .52 \\
BioCLIP~2 & TURTLE~\cite{gadetsky2024turtle} & .53 & .47 & .40 & .29 & .29 \\
\midrule
BioCLIP~2 & HDBSCAN (\textsc{H}), $K{\approx}97$ & .53 & .79 & .66 & .50 & .49 \\
BioCLIP~2 & Leiden, $K{\approx}146$ & .51 & .79 & .70 & .54 & .53 \\
\midrule
BioCLIP~2 & SimGCD~\cite{wen2023simgcd}, semi-sup. & .60 & .81 & .77 & .59 & .57 \\
\bottomrule
\end{tabular}%
}
\end{table}

\subsection{Image-only clustering on BIOSCAN-5M}
\label{sec:exp_image_only}

We first evaluate image-only clustering on BIOSCAN-5M, comparing our reference pipeline with a matched reimplementation of the published image baseline and alternative clustering methods. Table~\ref{tab:methods} provides a quantitative comparison across two feature sets, clustering methods, and class-count requirements using a common AMI evaluation protocol.

\noindent\textbf{Comparison with the published baseline.}
The BIOSCAN-5M baseline uses ImageNet-pretrained DINO ViT-B features~\cite{caron2021emerging}, UMAP, and Ward clustering~\cite{gharaee2024bioscan5m}. Its released implementation evaluates a single partition with $K=4{,}363$ clusters, corresponding to the number of species, across all taxonomic ranks. Our reimplementation reproduces the published DINO curves, read from the vector figures of its arXiv versions, within 0.003 under the corresponding AMI normalisation, while Table~\ref{tab:methods} reports all methods using arithmetic-normalised AMI. Under this matched clustering protocol, replacing DINO with BioCLIP~2 increases family AMI from 0.36 to 0.52 and genus AMI from 0.29 to 0.58. However, BioCLIP~2 was pretrained with taxonomic supervision that includes the evaluation specimens, so this comparison does not establish generalisation to unseen specimens.

Because the fixed species-level cluster count is not appropriate for all taxonomic ranks, we also evaluate Ward clustering using the true number of classes at each rank (oracle $K$). Under this setting, DINO achieves 0.51 family and 0.34 genus AMI, compared with 0.73 and 0.65 for BioCLIP~2. These oracle-$K$ results provide a separate diagnostic comparison rather than a fully label-free operating point.

\noindent\textbf{Reference clustering pipeline.}
Our reference pipeline, BioCLIP~2 with UMAP and HDBSCAN (protocol \textsc{H}), achieves 0.79 family, 0.66 genus, and 0.50 species AMI, with $97\pm2$ clusters and 6.8\% of specimens assigned as noise.
Protocol \textsc{H} does not require the number of classes to form its partition, although its configuration was informed by family-level performance.
As discussed in Sec.~\ref{sec:exp_setup}, the minimum cluster size of 100 constrains the granularity of the resulting partition, particularly at species level.

\begin{table}[!t]
\centering
\caption{\textbf{Sensitivity to HDBSCAN minimum cluster size and noise handling on BIOSCAN-5M.}
Species-level AMI and clustering efficiency $\eta$ for BioCLIP~2, averaged over five UMAP seeds.
$K$ excludes noise, and the noise fraction is measured before reassignment.
Noise specimens are either treated as an additional cluster or assigned to the cluster of their nearest clustered neighbour in UMAP space.
The reference configuration is in bold.}
\label{tab:hdbscan_sensitivity}
\resizebox{0.65\linewidth}{!}{%
\begin{tabular}{rcccccc}
\toprule
 & & & \multicolumn{2}{c}{Noise as one cluster}
     & \multicolumn{2}{c}{Noise reassigned} \\
\cmidrule(lr){4-5}
\cmidrule(lr){6-7}
Min. size & $K$ & Noise & AMI & $\eta$ & AMI & $\eta$ \\
\midrule
10  & 839 & 26\% & .45 & .57 & .60 & .73 \\
20  & 543 & 21\% & .48 & .58 & .59 & .70 \\
25  & 434 & 20\% & .48 & .57 & .58 & .69 \\
50  & 191 & 11\% & .50 & .55 & .54 & .61 \\
\textbf{100} & \textbf{97} & \textbf{7\%}
 & \textbf{.50} & \textbf{.53}
 & \textbf{.51} & \textbf{.55} \\
200 & 58 & 7\% & .49 & .49 & .49 & .50 \\
300 & 41 & 5\% & .47 & .47 & .48 & .47 \\
\bottomrule
\end{tabular}%
}
\end{table}

\noindent\textbf{Sensitivity to partition granularity and noise handling.}
Table~\ref{tab:hdbscan_sensitivity} reports a post-hoc sensitivity analysis varying the minimum cluster size while retaining the remaining reference configuration and the same five UMAP seeds.
Reducing the minimum size from 100 to 20 increases the number of clusters from 97 to 543 and the noise fraction from 7\% to 21\%.
Under AMI(global), species AMI decreases from 0.50 to 0.48, while clustering efficiency increases from 0.53 to 0.58.
After reassigning noise specimens without labels, species AMI and efficiency increase to 0.59 and 0.70, compared with 0.51 and 0.55 for the reference partition treated in the same way.
This finer partition improves species-level agreement at the expense of family-level agreement (0.72 versus 0.81 AMI).
Thus, the magnitude of the measured clustering limitation depends on partition granularity and noise handling, although substantial species-level headroom remains.

\begin{figure}[!t]
\centering
\includegraphics[width=\linewidth]{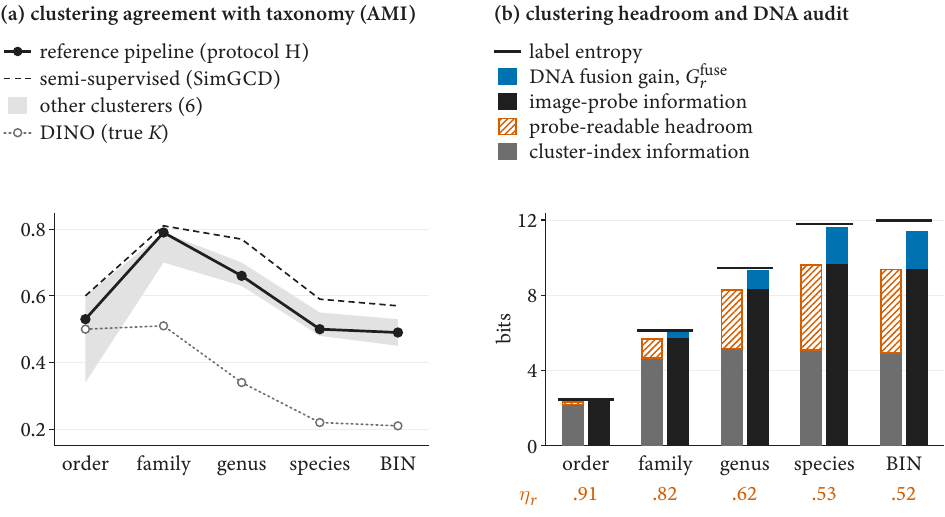}
\caption{\textbf{Clustering and representation limits both grow towards species on BIOSCAN-5M.}
(a) AMI by rank for the reference pipeline (protocol \textsc{H}: BioCLIP~2 $\rightarrow$ UMAP $\rightarrow$ HDBSCAN), semi-supervised SimGCD, six alternative partitions on BioCLIP~2 embeddings (grey range: oracle-$K$ Ward, k-means, TEMI and SCAN, oracle-level FINCH, and Leiden), and DINO with the true class count $K$ (Table~\ref{tab:methods}).
(b) Held-out information for BioCLIP~2 under protocol \textsc{H} (Table~\ref{tab:info}). 
Left bar: cluster-index information (grey) and probe-readable headroom up to the image probe (orange; the clustering limit). 
Right bar: image-probe information (black) and DNA fusion gain $G^{\mathrm{fuse}}_r$ (blue; the representation limit). Black ticks mark label entropy; $\eta_r$ is listed below each rank. 
The two bars are complementary diagnostics, not an additive decomposition.}
\label{fig:ami_ranks}\label{fig:budget}
\end{figure}


\noindent\textbf{Comparison with alternative clustering methods.}
To examine whether the decline towards finer taxonomic ranks persists across clustering methods, Figure~\ref{fig:ami_ranks}(a) compares the rank-wise performance profiles of representative configurations from Table~\ref{tab:methods}.
Although the reference HDBSCAN pipeline is competitive at family and genus, alternative partitions recover additional structure from the same frozen features.
In particular, Leiden produces approximately 146 clusters and improves genus AMI from 0.66 to 0.70 and species AMI from 0.50 to 0.54 without requiring the number of classes.
When lowering the Leiden resolution (0.20--0.26) to approximately match the number of clusters produced by protocol \textsc{H}, the gains shrink to 0.01 at both ranks and disappear when noise specimens in the reference partition are reassigned.
This suggests that partition granularity and noise handling account for much of the observed improvement.
Several oracle-$K$ and learned methods also improve species clustering, while SimGCD reaches 0.59 species AMI using partial label supervision and is therefore not directly comparable to the label-free methods.


These results establish a clustering-limited component of the species-level plateau: finer partitions and alternative clustering configurations recover additional taxonomic structure from the same representation.
Nevertheless, the decline towards species and BIN persists across the representative configurations in Figure~\ref{fig:ami_ranks}(a), while Table~\ref{tab:hdbscan_sensitivity} shows that substantial species-level clustering headroom remains even for finer partitions.
We next examine how much probe-readable taxonomic information the reference partition retains and whether paired DNA provides a predictive signal beyond the frozen image embedding.

\begin{table}[!t]
\centering
\caption{Rank-wise information audit for BioCLIP~2 under protocol \textsc{H} on BIOSCAN-5M. All information values are in bits; $\eta$ is dimensionless. $\widehat{I}_{\mathrm{cv}}(C;Y)$ is held-out cluster-index information, and $G^{\mathrm{fuse}}$ is the held-out gain from adding DNA to image. Fusion gains are mean $\pm$ standard deviation over three probe seeds; cluster information and $\eta$ are averaged over five UMAP seeds (standard deviations $\leq$ 0.021 bits and $\leq 0.005$, respectively).}
\label{tab:info}
\resizebox{0.8\linewidth}{!}{%
\begin{tabular}{lcccccc}
\toprule
Rank & $H(Y)$ & $\widehat{I}_{\phi}(Z^{\mathrm{img}};Y)$ & $\widehat{I}_{\phi}(Z^{\mathrm{dna}};Y)$ & $G^{\mathrm{fuse}}$ & $\widehat{I}_{\mathrm{cv}}(C;Y)$ & $\eta$ \\
\midrule
Order   & 2.46  & 2.39 & 2.45  & $0.06\pm.00$ & 2.19 & .91 \\
Family  & 6.13  & 5.72 & 6.08  & $0.35\pm.00$ & 4.67 & .82 \\
Genus   & 9.46  & 8.33 & 9.36  & $1.02\pm.01$ & 5.18 & .62 \\
Species & 11.80 & 9.64 & 11.65 & $1.99\pm.01$ & 5.09 & .53 \\
BIN     & 11.97 & 9.41 & 11.52 & $2.02\pm.00$ & 4.92 & .52 \\
\bottomrule
\end{tabular}%
}
\end{table}

\subsection{Clustering efficiency}
\label{sec:exp_eta}

We next examine how much taxonomic information the reference clustering partition retains relative to what can be read from the same frozen image representation. Table~\ref{tab:info} reports the held-out cluster-index information and clustering efficiency $\eta$ for BioCLIP~2 under protocol \textsc{H} (Sec.~\ref{sec:method_eta}).

At order and family, $\eta$ reaches 0.91 and 0.82, respectively, indicating that the partition captures taxonomic association comparable in scale to that accessible to the image probe. The ratio decreases to 0.62 at genus and 0.53 at species. 
At species, the held-out cluster-index estimate is 5.09 bits, compared with 9.64 bits from the image probe. This difference indicates substantial clustering headroom relative to the supervised probe; a finer partition recovers part of it (Sec.~\ref{sec:exp_image_only}).

Clustering efficiency and AMI capture different aspects of the partition. At order, for example, protocol \textsc{H} produces approximately 97 clusters for 27 taxonomic orders. Despite high efficiency ($\eta=0.91$), AMI is only 0.53, consistent with over-segmentation of the taxonomic classes. Thus, clustering efficiency describes the predictive association retained by cluster assignments relative to the image probe, whereas AMI evaluates agreement with the target taxonomy.

The held-out estimator also reduces the positive bias associated with plug-in cluster--label information. At species, a Miller--Madow-corrected plug-in estimate gives $\eta=0.58$, but still yields 0.19 under label permutation. In contrast, the held-out estimate is negative under permutation and 1.20 when the true species labels are used as the partition, providing a more informative reference for distinguishing meaningful association from estimator bias.

\subsection{DNA audit of the species-level plateau}
\label{sec:exp_dna_audit}

We next examine whether paired DNA provides taxonomic information beyond that accessible to the frozen image representation. 
Table~\ref{tab:info} reports the held-out late-fusion gain $G^{\mathrm{fuse}}$ for BioCLIP~2 across taxonomic ranks. Unlike clustering efficiency, the fusion gain is estimated directly from frozen embeddings and does not depend on the clustering configuration or the granularity of the resulting partition. 
The gain is small at order and family (0.06 and 0.35 bits, respectively), but increases to 1.02 bits at genus and 1.99 bits at species. Thus, adding DNA to the image embedding reduces held-out species-level prediction cross-entropy by approximately two bits under the evaluated probe protocol.

At species, the image probe exposes 9.64 bits of the 11.80-bit label entropy, whereas the DNA probe exposes 11.65 bits. The marginal difference between these estimates, $\Delta=2.01$ bits, is close to the late-fusion gain of 1.99 bits. However, $\Delta$ compares separate modality probes and does not independently establish information beyond the image embedding. The late-fusion gain provides the relevant evidence: paired DNA improves held-out prediction beyond the tested image representation and probe. It does not quantify an irreducible information gap in the underlying images.

Figure~\ref{fig:budget}(b) visualises how the two diagnostic limitations evolve across taxonomic ranks. While image-probe information and DNA fusion gain increase towards species, the held-out cluster-index information of the reference partition largely plateaus from genus onwards. The paired bars distinguish probe-readable structure not retained by the partition from the additional predictive gain obtained through DNA fusion. At species, these quantities are 4.55 and 1.99 bits, respectively. The former depends on the clustering configuration, whereas the latter does not. They are complementary diagnostics rather than an additive decomposition of information loss.


\noindent\textbf{Dependence on the image representation.}
Table~\ref{tab:encoders} extends the species-level audit to seven frozen image encoders. The late-fusion gain remains positive across all encoders, ranging from 1.10 to 4.77 bits. CLIBD, whose image tower was pretrained using paired BIOSCAN image--DNA data, exposes the most species-level information (10.54 bits) and has the smallest auxiliary gain (1.10 bits). BioCLIP~2 follows with 9.64 bits and a 1.99-bit gain. In contrast, general-purpose visual encoders expose less species-level information: DINOv2, for example, yields 8.15 bits and a 3.45-bit gain. 
A similar pattern is observed for DINOv3, which was not pretrained with explicit BIOSCAN taxonomic supervision: species-level AMI remains modest (0.42 with Leiden), while the DNA fusion gain reaches 2.94 bits. Thus, the coexistence of limited species-level clustering performance and substantial auxiliary gain is not specific to BioCLIP~2.

These differences show that the size of the DNA-audited gap depends strongly on the image representation. Because the DNA probe already predicts species labels with low held-out cross-entropy, the gains also share a common upper reference determined by barcode predictability. The results therefore characterise the limitations of different image embeddings relative to the same auxiliary signal, rather than providing independent estimates of missing visual information. The biological encoders also differ in their pretraining supervision, as described in Sec.~\ref{sec:exp_setup}.

\begin{table}[!t]
\centering
\caption{\textbf{Species-level DNA audit across frozen image encoders on BIOSCAN-5M.}
Image-probe information and late-fusion gain are in bits (three probe seeds). For Leiden and protocol \textsc{H}, $K$ denotes the mean number of clusters. Leiden AMI is reported at genus and species, while protocol \textsc{H} reports species AMI (five UMAP seeds). Gain standard deviations are at most 0.01 bits for the reported values; Leiden AMI standard deviations are at most 0.006. $^{\ast}$BioCLIP~2 received taxonomic supervision overlapping the evaluation specimens; CLIBD was pretrained using paired BIOSCAN image--DNA data.}
\label{tab:encoders}
\resizebox{0.85\linewidth}{!}{%
\begin{tabular}{lccccccc}
\toprule
 & \multicolumn{2}{c}{Species-level information audit}
 & \multicolumn{3}{c}{Leiden}
 & \multicolumn{2}{c}{Protocol \textsc{H}} \\
\cmidrule(lr){2-3}
\cmidrule(lr){4-6}
\cmidrule(lr){7-8}
Image encoder
 & $\widehat{I}_{\phi}(Z^{\mathrm{img}};Y)$
 & $G^{\mathrm{fuse}}$
 & $K$
 & \shortstack{Genus\\AMI}
 & \shortstack{Species\\AMI}
 & $K$
 & \shortstack{Species\\AMI} \\
\midrule
BioCLIP~2$^{\ast}$~\cite{gu2025bioclip2}
 & 9.64 & $1.99\pm.01$ & 146 & .70 & .54 & 97 & $.50\pm.00$ \\
CLIBD$^{\ast}$~\cite{gong2024clibd}
 & 10.54 & $1.10\pm.01$ & 227 & .72 & .58 & 124 & $.51\pm.00$ \\
DINOv3-L~\cite{simeoni2025dinov3}
 & 8.67 & $2.94\pm.01$ & 57 & .55 & .42 & 32 & $.34\pm.00$ \\
DINOv2-L~\cite{oquab2024dinov2}
 & 8.15 & $3.45\pm.01$ & 41 & .51 & .38 & 20 & $.33\pm.00$ \\
SigLIP-L~\cite{zhai2023siglip}
 & 7.25 & $4.33\pm.01$ & 35 & .44 & .34 & 26 & $.23\pm.10$ \\
MetaCLIP-L~\cite{xu2024metaclip}
 & 7.18 & $4.40\pm.01$ & 37 & .41 & .32 & 12 & $.15\pm.12$ \\
EVA02-B~\cite{fang2024eva02}
 & 6.82 & $4.77\pm.01$ & 33 & .39 & .30 & 2 & $.02\pm.00$ \\
\bottomrule
\end{tabular}%
}
\end{table}


\noindent\textbf{Clustering across representations.}
Table~\ref{tab:encoders} also reveals the sensitivity of the reference HDBSCAN configuration to encoder choice. For general-purpose visual encoders, protocol \textsc{H} produces between 2 and 32 clusters on average, with substantial variability in species-level AMI for SigLIP and MetaCLIP. This sensitivity is expected given differences in embedding geometry and the use of a fixed clustering configuration across encoders. Although encoder-specific hyperparameter tuning could improve HDBSCAN performance, Leiden is stable across seeds for every encoder (AMI standard deviation at most 0.006) and is therefore used for subsequent cross-encoder comparisons.

\noindent\textbf{DNA-only partitions.}
As a complementary clustering comparison, oracle-$K$ Ward partitions of mean-centred HyenaDNA embeddings achieve AMI of 0.34, 0.51, 0.69, and 0.90 from order to species, respectively. 
The corresponding oracle-$K$ Ward results for BioCLIP~2 in Table~\ref{tab:methods} are 0.59, 0.73, 0.65, and 0.53. DNA therefore produces substantially stronger species-level partitions in this setting, although this comparison is distinct from the held-out fusion audit and does not measure conditional information beyond image.

\subsection{Robustness of the DNA audit}
\label{sec:exp_robustness}

We evaluate whether the species-level DNA gain depends on the auxiliary representation, probe construction, barcode duplication, or taxonomic label provenance. Table~\ref{tab:robust_delta} summarises these controls for BioCLIP~2 and DINOv2.

\begin{table}[t]
\centering
\caption{\textbf{Robustness of the species-level DNA audit on BIOSCAN-5M.}
The upper section reports held-out late-fusion gain $G^{\mathrm{fuse}}$ in bits, averaged over three probe seeds unless otherwise indicated (standard deviations $\leq 0.02$). The lower section reports species-level image-probe information $\widehat{I}_{\phi}(Z^{\mathrm{img}};Y)$ in bits for alternative probe configurations (three seeds; the wider MLP has 4,096 hidden units, the nearest-neighbour estimate uses 15 neighbours). Unless specified, the auxiliary is HyenaDNA and evaluation uses random folds. $^{\dagger}$The fold and label-provenance controls use 45,965 specimens from species with at least two distinct barcodes; 42,704 remain after excluding copied labels. Barcode-grouped folds keep identical sequences together and are class-stratified. $^{\ddagger}$DINOv2 probe-schedule results use one seed.}
\label{tab:robust_delta}
\resizebox{0.8\linewidth}{!}{%
\begin{tabular}{llcc}
\toprule
Control & Setting & BioCLIP~2 & DINOv2-L \\
\midrule
\multicolumn{4}{l}{\textit{Late-fusion gain $G^{\mathrm{fuse}}$ (bits)}} \\
\addlinespace
Reference & HyenaDNA & 1.99 & 3.45 \\
DNA encoder & CLIBD DNA tower & 2.00 & 3.47 \\
\midrule
Null auxiliary & Permuted DNA & $-1.27$ & $-1.84$ \\
Null auxiliary & Gaussian noise & $-1.02$ & $-1.55$ \\
Other auxiliary & Collection locality & 0.41 & 0.86 \\
Other auxiliary & Specimen size & $-0.29$ & $-0.22$ \\
\midrule
Probe folds$^{\dagger}$ & Random / barcode-grouped & 1.94 / 2.32 & 3.37 / 3.89 \\
Label provenance & Non-placeholder species & 1.84 & -- \\
Label provenance$^{\dagger}$ & Excluding copied labels & 1.94 & 3.42 \\
\midrule
Class size & $\geq 5$ / $\geq 10$ / $\geq 20$ specimens
    & 1.70 / 1.29 / 0.97 & -- / 2.34 / -- \\
Probe schedule$^{\ddagger}$ & 20 / 40 / 80 epochs
    & 1.85 / 1.90 / 1.99 & 3.38 / 3.38 / 3.44 \\
\midrule
\multicolumn{4}{l}{\textit{Image-probe information $\widehat{I}_{\phi}(Z^{\mathrm{img}};Y)$ (bits)}} \\
\addlinespace
Reference probe & Shallow MLP & 9.64 & 8.15 \\
Probe family & Linear & 9.61 & 7.55 \\
Probe capacity & Wider MLP & 9.39 & 7.63 \\
Probe family & Nearest neighbour & 8.47 & 3.76 \\
\bottomrule
\end{tabular}%
}
\end{table}

\noindent\textbf{Auxiliary representations and controls.}
Replacing HyenaDNA with the CLIBD DNA tower changes the species-level gain by about 0.01 bits for both BioCLIP~2 and DINOv2 (0.009 and 0.012 before rounding), indicating that the result is consistent across the two DNA encoders tested. In contrast, permuting DNA sequences across specimens or replacing them with Gaussian noise produces negative gains. These controls show that the positive DNA gain is not reproduced by an uninformative auxiliary, although the negative values also indicate sensitivity to the optimisation of the fusion probe. Other specimen metadata provide smaller gains: collection locality contributes 0.41 bits beyond BioCLIP~2 and 0.86 bits beyond DINOv2, whereas specimen-size measurements do not improve held-out prediction.

\noindent\textbf{Barcode duplication and label provenance.}
Identical barcodes are common in BIOSCAN-5M, making random cross-validation potentially sensitive to repeated sequences. We therefore repeat the audit using barcode-grouped folds, ensuring that specimens with identical barcodes are not shared between probe training and evaluation. On the same subset of specimens, the BioCLIP~2 gain increases from 1.94 to 2.32 bits, while the DINOv2 gain increases from 3.37 to 3.89 bits. Thus, the species-level gain persists without exact-barcode overlap between folds.

We also examine the dependence on label provenance. Restricting the evaluation to species with non-placeholder names yields a BioCLIP~2 gain of 1.84 bits. Excluding specimens whose labels were copied from another specimen with the same barcode leaves the gain at 1.94 bits on the corresponding subset. These controls reduce concerns about label construction, although they do not establish that species assignments are independent of barcode evidence.

\noindent\textbf{Class-size dependence.}
The audit is sensitive to the number of specimens available per species. For BioCLIP~2, the held-out gain decreases from 1.99 bits on the full evaluation set to 1.70, 1.29, and 0.97 bits when retaining only species with at least 5, 10, and 20 specimens, respectively. 
The gain is also unevenly distributed across specimens: its median is 0.31 bits, while the highest-gain tenth accounts for 53\% of the total. These results show that rare and poorly sampled classes contribute substantially to the overall gap. The approximately two-bit species gain should therefore be interpreted in the context of the benchmark's sparse sampling (on average eleven specimens per species, at most 33) rather than as a fixed difference between image and DNA information.

\noindent\textbf{Probe sensitivity.}
Probe configuration affects both fusion gain and probe-readable information. For BioCLIP~2, increasing training from 20 to 80 epochs raises $G^{\mathrm{fuse}}$ from 1.85 to 1.99 bits. The shallow MLP provides the highest species-level information estimate for both encoders (Table~\ref{tab:robust_delta}). 
For DINOv2, the linear and nearest-neighbour probes read 0.6 and 4.4 bits less, respectively; the latter estimate also depends on neighbourhood size and sparse class sampling. Label permutation produces a negative information estimate, while increasing probe training data improves it. These findings reinforce the dependence of the diagnostic on the probe and evaluation protocol.


Overall, the species-level DNA gain persists across the tested auxiliary encoders, barcode-grouped folds, and label-provenance controls. Its magnitude nevertheless depends on class composition and probe specification, motivating the additional dataset evaluations in Sec.~\ref{sec:exp_additional}.

\subsection{Validation beyond the evaluation split}
\label{sec:exp_additional}

We further evaluate the stability of the diagnostic on disjoint BIOSCAN specimens and its applicability to other datasets.

\noindent\textbf{Disjoint BIOSCAN specimens.}
We repeat the audit on the BIOSCAN-5M key-unseen partition, which contains specimens disjoint from the main evaluation split, but includes overlapping species. Restricting both sets to species with at least ten specimens gives 33,805 specimens from 463 species in key-unseen and 33,387 from 1,732 species in the evaluation split. 
To assess the stability of the diagnostic across these subsets, we set pragmatic agreement tolerances of 0.05 for clustering efficiency and 0.15 bits for fusion gain before the run. Neither is met at species level. 
For BioCLIP~2, key-unseen achieves higher AMI (0.80 versus 0.56) and efficiency (0.76 versus 0.54; 0.77 versus 0.57 with the plug-in ratio for which the tolerance was written), but a smaller DNA gain (0.54 versus 1.29 bits).


These differences coincide with substantial variation in class composition: the ten largest key-unseen species account for 44\% of specimens, whereas no evaluation species has more than 33 specimens. 
When matching specimen counts for the 265 species with at least ten specimens in both key-unseen and test-unseen (4,709 specimens per set), the results closely agree: AMI is 0.59 in both, efficiency is 0.52 versus 0.51, and fusion gain is 0.88 versus 0.89 bits. This matched analysis was defined after inspecting class composition. The findings indicate that the diagnostic is sensitive to class distribution, while comparable populations yield similar estimates. They do not establish generalisation to unseen species.

\noindent\textbf{A second dataset and auxiliary signal.}
We also evaluate CUB-200-2011 using human-annotated visual attributes as the auxiliary signal. Unlike DNA, these attributes describe the same images used to extract visual embeddings, allowing us to test whether explicit visual annotations provide additional predictive information.

\begin{table}[t]
\centering
\caption{\textbf{InfoTaxa audit on CUB-200-2011 at species level.}
The species-label entropy is 7.65 bits, and the attribute-only probe exposes 4.47 bits. Late-fusion gains (bits) are evaluated using attributes from the same image (own), randomly permuted attributes, or attributes from another image of the same species. 
AMI is reported for HDBSCAN and Leiden (five UMAP seeds); $\eta$ uses the HDBSCAN partition. 
ResNet-18 (ImageNet-1k weights) is added as a weak encoder; ImageNet-1k includes bird classes, and its image overlap with CUB was not measured.
$^{\ast}$All 200 CUB species occur in the BioCLIP~2 training catalogue, but exact evaluation-image overlap is not established.}
\label{tab:cub}
\resizebox{0.8\linewidth}{!}{%
\begin{tabular}{lccccccc}
\toprule
 & & \multicolumn{3}{c}{Fusion gain}
 & \multicolumn{2}{c}{AMI} & \\
Image encoder
 & $\widehat{I}_{\phi}(Z^{\mathrm{img}};Y)$
 & Own
 & Permuted
 & Same species
 & HDBSCAN
 & Leiden
 & $\eta$ \\
\midrule
BioCLIP~2$^{\ast}$
 & 7.18 & $-0.03$ & $-0.02$ & 0.09 & .92 & .93 & .97 \\
DINOv3-L
 & 7.16 & $-0.05$ & $-0.07$ & 0.09 & .87 & .89 & .92 \\
DINOv2-L
 & 7.12 & $-0.06$ & $-0.13$ & 0.07 & .85 & .87 & .89 \\
MetaCLIP-L
 & 6.98 & $-0.03$ & $-0.11$ & 0.15 & .78 & .81 & .80 \\
SigLIP-L
 & 6.97 & $-0.03$ & $-0.12$ & 0.14 & .80 & .82 & .82 \\
EVA02-B
 & 6.91 & $-0.04$ & $-0.17$ & 0.14 & .74 & .76 & .75 \\
ResNet-18
 & 5.90 & $0.39$ & $-0.36$ & 0.68 & .47 & .54 & .49 \\
\bottomrule
\end{tabular}%
}
\end{table}

Table~\ref{tab:cub} shows a contrasting outcome to BIOSCAN-5M. Across six foundation-model image encoders, the image probes expose 6.91--7.18 of the 7.65 bits of species-label entropy. For these six, adding attributes from the same image provides no measurable held-out gain ($-0.06$ to $-0.03$ bits); the permuted-attribute null is itself negative ($-0.17$ to $-0.02$ bits), and signs change with the probe's batch size, so differences of about 0.05 bits are not resolved. Attributes from another image of the same species provide small positive gains (0.07--0.15 bits), but this control uses ground-truth species membership and is not a deployment setting.

For BioCLIP~2, HDBSCAN achieves 0.92 AMI and $\eta=0.97$, indicating that the partition retains most of the information accessible to the image probe. Across foundation-model encoders, probe information and clustering performance generally follow similar trends. Unlike DNA barcodes on BIOSCAN-5M, CUB attributes are derived from the same images used to construct the visual embeddings rather than from an independent measurement. The limited additional predictive value of these attributes is consistent with much of their label-relevant information already being accessible to the tested image probes. This does not establish that the frozen representations capture all species-relevant visual information.

In contrast, an ImageNet-pretrained ResNet-18 whose probe exposes 5.90 bits benefits from the same attributes ($0.39\pm0.02$ bits over three probe seeds), while permuted attributes yield a negative gain. Its clustering efficiency is also lower ($\eta=0.49$), although 33\% of images are assigned as noise. These results demonstrate that image-derived attributes can provide additional predictive value beyond a frozen visual representation.


\noindent\textbf{Probe-only evaluations.}
On NEON carabid beetles, BioCLIP~2 and DINOv2 expose 0.96 and 0.95 of species-label entropy across 47 species. On FishNet, the fraction exposed by BioCLIP~2 decreases from 0.87 at order to 0.77 at genus, but remains between 0.87 and 0.89 across ranks after restricting classes to at least ten images. These findings suggest that class size contributes to the apparent fine-rank decline. Without paired auxiliary signals, these datasets provide representation-level checks rather than full InfoTaxa audits.

\subsection{Acting on the diagnosis}
\label{sec:exp_use}

\begin{figure}[!t]
\centering
\includegraphics[width=\linewidth]{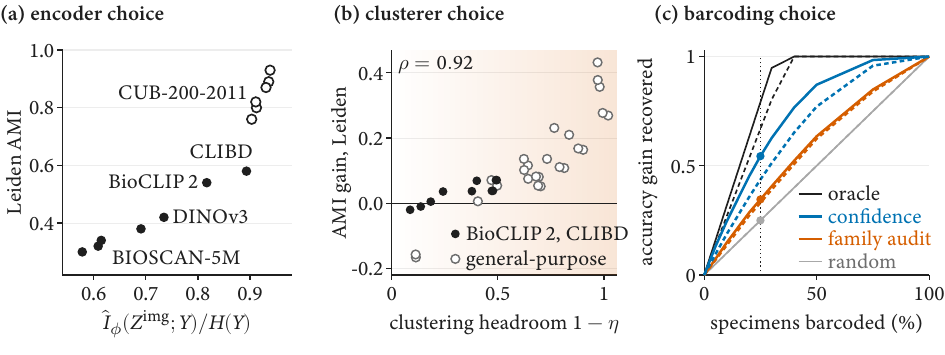}
\caption{\textbf{Acting on the diagnosis.}
(a) Encoder choice: image-probe information, normalised by label entropy, against species-level Leiden AMI (filled: BIOSCAN-5M; open: the six foundation-model encoders on CUB-200-2011).
(b) Clusterer choice: clustering headroom $1-\eta$ against the AMI gain obtained by replacing HDBSCAN with Leiden, across 35 encoder--rank combinations on BIOSCAN-5M.
(c) Barcoding choice: fraction of the available image-to-DNA gain in species accuracy recovered as the sequencing budget increases. Solid and dashed lines indicate random and barcode-grouped probe folds, respectively; markers indicate a 25\% sequencing budget.
}
\label{fig:use}
\end{figure}

We examine three potential uses of the diagnostic: encoder selection, assessing clustering headroom, and selective DNA barcoding. Figure~\ref{fig:use} summarises the results. The evaluation criteria were defined before the corresponding analyses, although the underlying probe and clustering results were already available when specifying the encoder-selection criteria. The thresholds are pragmatic evaluation targets rather than established standards.


\noindent\textbf{Selecting an image encoder.}
We test whether probe-readable information can guide encoder selection. As operational success criteria, we require a Spearman correlation of at least 0.8 with protocol-\textsc{H} AMI at genus and species, and a selected encoder within 0.01 AMI of the best. These criteria assess both agreement in encoder rankings and the performance of the selected encoder. Figure~\ref{fig:use}(a) illustrates the relationship between species-level probe-readable information and Leiden AMI across both datasets. On BIOSCAN-5M, the probe estimates rank all seven encoders consistently with Leiden AMI at genus and species, a secondary target reported first because protocol \textsc{H} is unstable on general-purpose encoders. 
Under protocol \textsc{H}, correlations remain high (0.96 and 1.00), but the selection criterion fails at genus: CLIBD is selected while BioCLIP~2 achieves 0.017 higher AMI, a difference that would satisfy a tolerance of 0.02 rather than 0.01. On CUB-200-2011, correlations with HDBSCAN AMI across the six foundation-model encoders reach 1.00 at genus and 0.94 at species.

Using only 2,000 labelled BIOSCAN specimens, the probe selects CLIBD in all five samples at genus and species, consistent with the full-data estimates. Consequently, the genus-level selection criterion remains unmet under protocol \textsc{H}. Species-level information estimates are negative in this setting, and selection therefore relies on relative held-out cross-entropy.
A label-free silhouette criterion correlates less strongly with Leiden AMI (0.57), while the number of HDBSCAN clusters provides a competitive correlation (0.96). Thus, probe estimates support fine-rank encoder comparison, but their advantage over simpler selection criteria is not established.

\noindent\textbf{Deciding whether to change the clusterer.}
We assess whether $1-\eta$ predicts improvement from alternative clustering methods across 35 encoder--rank combinations using Spearman rank correlation, with a target of at least 0.5. The observed correlations are 0.78 for the better of Leiden and oracle-level FINCH, and 0.92 for Leiden alone (Figure~\ref{fig:use}(b)). The criterion predates the held-out definition of $\eta$; with the original plug-in estimator, the first correlation was 0.80. Leiden improves all 31 combinations with $\eta\leq0.85$, but none of the four above this threshold (all at order). This threshold was identified from the results. For BioCLIP~2 at species, $\eta=0.53$ and Leiden improves AMI by 0.04. However, $1-\mathrm{AMI}$ of the original partition predicts these improvements similarly well (correlations of 0.80 and 0.87, respectively). Clustering efficiency therefore identifies potential headroom but does not demonstrate a clear advantage over AMI for selecting another clustering algorithm.


\noindent\textbf{Choosing specimens for DNA barcoding.}
We simulate selective sequencing using cross-validated image and DNA probes, where selected specimens receive the DNA probe's species prediction. Routing scores use only the held-out specimen's image and probes fitted on the training folds, including the family-level gap (inner cross-validation); only the oracle sees held-out labels and DNA predictions. Figure~\ref{fig:use}(c) compares routing strategies across sequencing budgets. At a 25\% sequencing budget, routing by the audited image--DNA gap of the predicted family recovers 0.35 of the available species-accuracy gain, compared with 0.25 for random selection. Although this improves upon random selection, it falls short of our operational success criterion of recovering twice the gain achieved by random selection. 
A post-hoc optimised ordering by predicted family recovers only 0.36 of the available gain, suggesting that the limitation is partly associated with the granularity of family-level routing.

Routing by low image-probe confidence performs better, recovering 0.54 of the available gain. Species accuracy increases from 0.67 to 0.84, compared with 0.99 under full sequencing. With barcode-grouped folds, confidence routing recovers 0.44 of the gain, increasing accuracy from 0.54 to 0.70, compared with 0.91 under full sequencing.

These results indicate that the DNA audit identifies taxonomic levels where sequencing is valuable, but family-level routing provides limited discrimination between individual specimens and does not outperform standard confidence-based selection. The evaluation remains a supervised simulation rather than a validated field-sequencing strategy.

%% file: sec/5_discussion_limitations.tex
\section{Discussion and limitations}
\label{sec:discussion}

Our findings indicate that species-level image-only clustering on BIOSCAN-5M is both clustering-limited and representation-limited under the evaluated protocol. Low clustering efficiency reveals probe-readable structure not retained by the reference partition, while the positive DNA fusion gain identifies additional predictive signal beyond the frozen image representation. 
Sensitivity analyses show that partition granularity and noise handling account for part of the clustering limitation, while the DNA fusion gain remains independent of the partition.
These measurements are complementary diagnostics rather than an additive decomposition of information loss.

The practical implications are more qualified. Encoder choice has a substantial effect on fine-grained clustering, and probe estimates correlate strongly with encoder performance. However, clustering efficiency does not predict the benefit of changing the clusterer substantially better than AMI alone. Similarly, family-level audit-based routing underperforms standard image-probe confidence for selective barcoding. Thus, \method{} helps characterise potential limitations but does not necessarily identify the optimal intervention.

Several factors constrain the interpretation. Clustering efficiency and fusion gain depend on probe configuration and class composition, while probe-readable information also depends on labelled sample size. Their values differ across disjoint BIOSCAN subsets but agree closely after matching class distributions. Species labels may be influenced by barcode evidence, and BioCLIP~2 received taxonomic supervision overlapping the evaluation specimens. The reference clustering configuration was selected using family-level performance. Furthermore, fusion gain measures held-out predictive improvement, not information proven absent from the original images.

Finally, the full audit with an independently measured auxiliary modality is evaluated only on BIOSCAN-5M. 
On CUB-200-2011, image-derived attributes provide negligible gain for foundation-model encoders but positive gain for ResNet-18, further illustrating the dependence of the diagnosis on the representation.
Evaluation on additional datasets with independent auxiliary measurements and controlled pretraining exposure remains necessary.

%% file: sec/6_conclusion.tex
\section{Conclusion}
\label{sec:conclusion}

We presented \method{}, an information-calibrated diagnostic for label-free clustering of frozen visual representations that combines partition quality, clustering efficiency, and held-out auxiliary gain. 
On BIOSCAN-5M, the analysis identifies both probe-readable structure that the reference partition does not recover and additional DNA-predictive signal at the species level. 
On CUB-200-2011, image-derived attributes provide no measurable additional gain for six foundation-model encoders, but improve prediction for ResNet-18.
The estimated diagnostic quantities depend on probe configuration and class composition, while pretraining overlap limits generalisation claims. 
\method{} helps identify where fine-grained clustering is limited and whether further investigation should focus on the partition or the representation. Translating these diagnostics into effective intervention strategies remains an open research direction.

%% file: main.bib
@String(CVPR= {IEEE Conf. Comput. Vis. Pattern Recog.})

@String(ICCV= {Int. Conf. Comput. Vis.})

@String(ECCV= {Eur. Conf. Comput. Vis.})

@String(ICLR = {Int. Conf. Learn. Represent.})

@String(CVPR  = {CVPR})

@String(ICCV  = {ICCV})

@String(ECCV  = {ECCV})

@String(ICLR  = {ICLR})

@article{hebert2003barcodes,
  author    = {Hebert, Paul D. N. and Cywinska, Alina and Ball, Shelley L. and deWaard, Jeremy R.},
  title     = {Biological Identifications through {DNA} Barcodes},
  journal   = {Proceedings of the Royal Society B: Biological Sciences},
  volume    = {270},
  number    = {1512},
  pages     = {313--321},
  year      = {2003}
}

@article{bickford2007cryptic,
  author    = {Bickford, David and Lohman, David J. and Sodhi, Navjot S. and Ng, Peter K. L. and Meier, Rudolf and Winker, Kevin and Ingram, Kristen K. and Das, Indraneil},
  title     = {Cryptic Species as a Window on Diversity and Conservation},
  journal   = {Trends in Ecology \& Evolution},
  volume    = {22},
  number    = {3},
  pages     = {148--155},
  year      = {2007}
}

@article{gharaee2024bioscan5m,
  title={BIOSCAN-5M: A multimodal dataset for insect biodiversity},
  author={Gharaee, Zahra and Lowe, Scott C and Gong, ZeMing and Arias, Pablo M and Pellegrino, Nicholas and Wang, Austin T and Haurum, Joakim B and Zarubiieva, Iuliia and Kari, Lila and Steinke, Dirk and others},
  journal={Advances in Neural Information Processing Systems},
  volume={37},
  pages={36285--36313},
  year={2024}
}

@inproceedings{stevens2024bioclip,
  title={Bioclip: A vision foundation model for the tree of life},
  author={Stevens, Samuel and Wu, Jiaman and Thompson, Matthew J and Campolongo, Elizabeth G and Song, Chan Hee and Carlyn, David Edward and Dong, Li and Dahdul, Wasila M and Stewart, Charles and Berger-Wolf, Tanya and others},
  booktitle={Proceedings of the IEEE/CVF Conference on Computer Vision and Pattern Recognition (CVPR)},
  pages={19412--19424},
  year={2024}
}

@article{gu2025bioclip2,
  title={Bioclip 2: Emergent properties from scaling hierarchical contrastive learning},
  author={Gu, Jianyang and Stevens, Sam and Campolongo, Elizabeth and Thompson, Matthew and Zhang, Net and Wu, Jiaman and Kopanev, Andrei and Mai, Zheda and White, Alexander and Balhoff, James and others},
  journal={Advances in Neural Information Processing Systems},
  volume={38},
  pages={102778--102811},
  year={2025}
}

@article{oquab2024dinov2,
  title={Dinov2: Learning robust visual features without supervision},
  author={Oquab, Maxime and Darcet, Timoth{\'e}e and Moutakanni, Th{\'e}o and Vo, Huy and Szafraniec, Marc and Khalidov, Vasil and Fernandez, Pierre and Haziza, Daniel and Massa, Francisco and El-Nouby, Alaaeldin and others},
  journal={Transactions on Machine Learning Research},
  year={2024}
}

@inproceedings{nguyen2024hyenadna,
  title={Hyenadna: Long-range genomic sequence modeling at single nucleotide resolution},
  author={Nguyen, Eric and Poli, Michael and Faizi, Marjan and Thomas, Armin and Wornow, Michael and Birch-Sykes, Callum and Massaroli, Stefano and Patel, Aman and Rabideau, Clayton and Bengio, Yoshua and others},
  booktitle={Advances in neural information processing systems},
  volume={36},
  pages={43177--43201},
  year={2023}
}

@inproceedings{van2020scan,
  title={Scan: Learning to classify images without labels},
  author={Van Gansbeke, Wouter and Vandenhende, Simon and Georgoulis, Stamatios and Proesmans, Marc and Van Gool, Luc},
  booktitle={European conference on computer vision},
  pages={268--285},
  year={2020},
  organization={Springer}
}

@inproceedings{wen2023simgcd,
  author    = {Wen, Xin and Zhao, Bingchen and Qi, Xiaojuan},
  title     = {Parametric Classification for Generalized Category Discovery: A Baseline Study},
  booktitle = {Proceedings of the IEEE/CVF International Conference on Computer Vision (ICCV)},
  year      = {2023}
}

@inproceedings{vaze2022gcd,
  author    = {Vaze, Sagar and Han, Kai and Vedaldi, Andrea and Zisserman, Andrew},
  title     = {Generalized Category Discovery},
  booktitle = {Proceedings of the IEEE/CVF Conference on Computer Vision and Pattern Recognition (CVPR)},
  pages     = {7492--7501},
  year      = {2022}
}

@inproceedings{van2018inaturalist,
  author    = {Van Horn, Grant and Mac Aodha, Oisin and Song, Yang and Cui, Yin and Sun, Chen and Shepard, Alex and Adam, Hartwig and Perona, Pietro and Belongie, Serge},
  title     = {The iNaturalist Species Classification and Detection Dataset},
  booktitle = {Proceedings of the IEEE Conference on Computer Vision and Pattern Recognition (CVPR)},
  pages     = {8769--8778},
  year      = {2018}
}

@inproceedings{gong2024clibd,
  title={CLIBD: Bridging vision and genomics for biodiversity monitoring at scale},
  author={Gong, ZeMing and Wang, Austin and Huo, Xiaoliang and Haurum, Joakim Bruslund and Lowe, Scott C and Taylor, Graham W and Chang, Angel},
  booktitle={International Conference on Learning Representations},
  pages={92306--92336},
  year={2025}
}

@inproceedings{mcallester2020formallimitations,
  title={Formal limitations on the measurement of mutual information},
  author={McAllester, David and Stratos, Karl},
  booktitle={International Conference on Artificial Intelligence and Statistics},
  pages={875--884},
  year={2020},
  organization={PMLR}
}

@inproceedings{tishby1999information,
  author    = {Tishby, Naftali and Pereira, Fernando C. and Bialek, William},
  title     = {The Information Bottleneck Method},
  booktitle = {Proceedings of the 37th Annual Allerton Conference on Communication, Control, and Computing},
  pages     = {368--377},
  year      = {1999}
}

@incollection{miller1955note,
  author    = {Miller, George A.},
  title     = {Note on the Bias of Information Estimates},
  booktitle = {Information Theory in Psychology: Problems and Methods},
  editor    = {Quastler, Henry},
  publisher = {Free Press},
  address   = {Glencoe, IL},
  pages     = {95--100},
  year      = {1955}
}

@article{mcinnes2018umap,
  title={Umap: Uniform manifold approximation and projection for dimension reduction},
  author={McInnes, Leland and Healy, John and Melville, James},
  journal={arXiv preprint arXiv:1802.03426},
  year={2018}
}

@article{mcinnes2017hdbscan,
  author  = {McInnes, Leland and Healy, John and Astels, Steve},
  title   = {{HDBSCAN}: Hierarchical Density Based Clustering},
  journal = {Journal of Open Source Software},
  volume  = {2},
  number  = {11},
  pages   = {205},
  year    = {2017}
}

@article{vinh2010information,
  author  = {Vinh, Nguyen Xuan and Epps, Julien and Bailey, James},
  title   = {Information Theoretic Measures for Clusterings Comparison: Variants, Properties, Normalization and Correction for Chance},
  journal = {Journal of Machine Learning Research},
  volume  = {11},
  pages   = {2837--2854},
  year    = {2010}
}

@inproceedings{zhai2023siglip,
  title={Sigmoid loss for language image pre-training},
  author={Zhai, Xiaohua and Mustafa, Basil and Kolesnikov, Alexander and Beyer, Lucas},
  booktitle={2023 IEEE/CVF International Conference on Computer Vision (ICCV)},
  pages={11941--11952},
  year={2023},
  organization={IEEE}
}

@article{fang2024eva02,
  title={Eva-02: A visual representation for neon genesis},
  author={Fang, Yuxin and Sun, Quan and Wang, Xinggang and Huang, Tiejun and Wang, Xinlong and Cao, Yue},
  journal={Image and Vision Computing},
  volume={149},
  pages={105171},
  year={2024},
}

@inproceedings{sarfraz2019finch,
  title={Efficient parameter-free clustering using first neighbor relations},
  author={Sarfraz, Saquib and Sharma, Vivek and Stiefelhagen, Rainer},
  booktitle={2019 IEEE/CVF Conference on Computer Vision and Pattern Recognition (CVPR)},
  pages={8926--8935},
  year={2019},
  organization={IEEE}
}

@article{traag2019leiden,
  author  = {Traag, Vincent A. and Waltman, Ludo and van Eck, Nees Jan},
  title   = {From Louvain to Leiden: Guaranteeing Well-Connected Communities},
  journal = {Scientific Reports},
  volume  = {9},
  number  = {1},
  pages   = {5233},
  year    = {2019}
}

@inproceedings{khan2023fishnet,
  title={Fishnet: A large-scale dataset and benchmark for fish recognition, detection, and functional trait prediction},
  author={Khan, Faizan Farooq and Li, Xiang and Temple, Andrew J and Elhoseiny, Mohamed},
  booktitle={Proceedings of the IEEE/CVF International Conference on Computer Vision (ICCV)},
  pages={20496--20506},
  year={2023}
}

@inproceedings{adaloglou2023temi,
  title={{Exploring the Limits of Deep Image Clustering using Pretrained Models}},
  author={Adaloglou, Nikolas and Michels, Felix and Kalisch, Hamza and Kollmann, Markus},
  booktitle={Proceedings of the British Machine Vision Conference},
  year={2023}
}

@inproceedings{gadetsky2024turtle,
  title={{Let Go of Your Labels with Unsupervised Transfer}},
  author={Gadetsky, Artyom and Jiang, Yulun and Brbic, Maria},
  booktitle={Proceedings of the 41st International Conference on Machine Learning},
  volume={235},
  pages={14382--14407},
  year={2024}
}

@inproceedings{xu2024metaclip,
  title     = {Demystifying {CLIP} Data},
  author    = {Xu, Hu and Xie, Saining and Tan, Xiaoqing Ellen and Huang, Po-Yao and Howes, Russell and Sharma, Vasu and Li, Shang-Wen and Ghosh, Gargi and Zettlemoyer, Luke and Feichtenhofer, Christoph},
  booktitle = {International Conference on Learning Representations (ICLR)},
  year      = {2024}
}

@article{simeoni2025dinov3,
  title   = {{DINOv3}},
  author={Sim{\'e}oni, Oriane and Vo, Huy V and Seitzer, Maximilian and Baldassarre, Federico and Oquab, Maxime and Jose, Cijo and Khalidov, Vasil and Szafraniec, Marc and Yi, Seungeun and Ramamonjisoa, Micha{\"e}l and others},
  journal = {arXiv preprint arXiv:2508.10104},
  year    = {2025}
}

@article{lowe2024empirical,
title = "An Empirical Study into Clustering of Unseen Datasets with Self-Supervised Encoders",
author={Lowe, Scott C. and Haurum, Joakim Bruslund
          and Oore, Sageev and Moeslund, Thomas B.
          and Taylor, Graham W.},
year = "2026",
month = aug,
language = "English",
volume = "2026-August",
journal = "Transactions on Machine Learning Research",
}

@inproceedings{cole2022contrastive,
  author    = {Cole, Elijah and Yang, Xuan and Wilber, Kimberly and Mac Aodha, Oisin and Belongie, Serge},
  title     = {When Does Contrastive Visual Representation Learning Work?},
  booktitle = {Proceedings of the IEEE/CVF Conference on Computer Vision and Pattern Recognition (CVPR)},
  pages     = {14755--14764},
  year      = {2022},
    organization={IEEE}

}

@article{pastore2023efficient,
  author  = {Pastore, Vito Paolo and Ciranni, Massimiliano and Bianco, Simone and Fung, Jennifer Carol and Murino, Vittorio and Odone, Francesca},
  title   = {Efficient unsupervised learning of biological images with compressed deep features},
  journal = {Image and Vision Computing},
  volume  = {137},
  pages   = {104764},
  year    = {2023},
}

@article{ciranni2024plankton,
  author  = {Ciranni, Massimiliano and Murino, Vittorio and Odone, Francesca and Pastore, Vito Paolo},
  title   = {Computer vision and deep learning meet plankton: Milestones and future directions},
  journal = {Image and Vision Computing},
  volume  = {143},
  pages   = {104934},
  year    = {2024},
}

@inproceedings{laina2022measuring,
  author    = {Laina, Iro and Asano, Yuki M. and Vedaldi, Andrea},
  title     = {Measuring the Interpretability of Unsupervised Representations via Quantized Reverse Probing},
  booktitle = {International Conference on Learning Representations (ICLR)},
  year      = {2022},
}

@article{he2025categorydiscovery,
  author  = {He, Zhenqi and Liu, Yuanpei and Han, Kai},
  title   = {Category Discovery: An Open-World Perspective},
  journal = {arXiv preprint arXiv:2509.22542},
  year    = {2025}
}

@inproceedings{zhao2023gpc,
  author    = {Zhao, Bingchen and Wen, Xin and Han, Kai},
  title     = {Learning Semi-supervised {Gaussian} Mixture Models for Generalized Category Discovery},
  booktitle = {Proceedings of the IEEE/CVF International Conference on Computer Vision (ICCV)},
  pages     = {16623--16633},
  year      = {2023}
}

@article{ratnasingham2013bin,
  author  = {Ratnasingham, Sujeevan and Hebert, Paul D. N.},
  title   = {A {DNA}-Based Registry for All Animal Species: The {Barcode Index Number} ({BIN}) System},
  journal = {PLoS ONE},
  volume  = {8},
  number  = {7},
  pages   = {e66213},
  year    = {2013},
}

@article{meier2006diptera,
  author  = {Meier, Rudolf and Shiyang, Kwong and Vaidya, Gaurav and Ng, Peter K. L.},
  title   = {{DNA} Barcoding and Taxonomy in {Diptera}: A Tale of High Intraspecific Variability and Low Identification Success},
  journal = {Systematic Biology},
  volume  = {55},
  number  = {5},
  pages   = {715--728},
  year    = {2006},
}

@inproceedings{badirli2021dna,
  author    = {Badirli, Sarkhan and Akata, Zeynep and Mohler, George and Picard, Christine and Dundar, Mehmet M.},
  title     = {Fine-Grained Zero-Shot Learning with {DNA} as Side Information},
  booktitle = {Advances in Neural Information Processing Systems},
  volume    = {34},
  pages     = {19352--19362},
  year      = {2021}
}

@article{badirli2023unknown,
  author  = {Badirli, Sarkhan and Picard, Christine Johanna and Mohler, George and Richert, Frannie and Akata, Zeynep and Dundar, Murat},
  title   = {Classifying the unknown: Insect identification with deep hierarchical {Bayesian} learning},
  journal = {Methods in Ecology and Evolution},
  volume  = {14},
  number  = {6},
  pages   = {1515--1530},
  year    = {2023}
}

@article{wuhrl2022diversityscanner,
  author  = {W{\"u}hrl, Lorenz and Pylatiuk, Christian and Giersch, Matthias and Lapp, Florian and von Rintelen, Thomas and Balke, Michael and Schmidt, Stefan and Cerretti, Pierfilippo and Meier, Rudolf},
  title   = {{DiversityScanner}: Robotic handling of small invertebrates with machine learning methods},
  journal = {Molecular Ecology Resources},
  volume  = {22},
  number  = {4},
  pages   = {1626--1638},
  year    = {2022}
}

@inproceedings{pimentel2020information,
  author    = {Pimentel, Tiago and Valvoda, Josef and Hall Maudslay, Rowan and Zmigrod, Ran and Williams, Adina and Cotterell, Ryan},
  title     = {Information-Theoretic Probing for Linguistic Structure},
  booktitle = {Proceedings of the 58th Annual Meeting of the Association for Computational Linguistics},
  pages     = {4609--4622},
  year      = {2020}
}

@inproceedings{voita2020information,
  author    = {Voita, Elena and Titov, Ivan},
  title     = {Information-Theoretic Probing with Minimum Description Length},
  booktitle = {Proceedings of the 2020 Conference on Empirical Methods in Natural Language Processing (EMNLP)},
  pages     = {183--196},
  year      = {2020}
}

@inproceedings{xu2020usable,
  author    = {Xu, Yilun and Zhao, Shengjia and Song, Jiaming and Stewart, Russell and Ermon, Stefano},
  title     = {A Theory of Usable Information under Computational Constraints},
  booktitle = {International Conference on Learning Representations (ICLR)},
  year      = {2020}
}

@inproceedings{hewitt2021conditional,
  author    = {Hewitt, John and Ethayarajh, Kawin and Liang, Percy and Manning, Christopher D.},
  title     = {Conditional probing: measuring usable information beyond a baseline},
  booktitle = {Proceedings of the 2021 Conference on Empirical Methods in Natural Language Processing (EMNLP)},
  pages     = {1626--1639},
  year      = {2021}
}

@article{romano2016adjusting,
  author  = {Romano, Simone and Vinh, Nguyen Xuan and Bailey, James and Verspoor, Karin},
  title   = {Adjusting for Chance Clustering Comparison Measures},
  journal = {Journal of Machine Learning Research},
  volume  = {17},
  number  = {134},
  pages   = {1--32},
  year    = {2016}
}

@inproceedings{slonim1999agglomerative,
  author    = {Slonim, Noam and Tishby, Naftali},
  title     = {Agglomerative Information Bottleneck},
  booktitle = {Advances in Neural Information Processing Systems},
  volume    = {12},
  pages     = {617--623},
  year      = {1999}
}

@article{ye2024r2trans,
  author  = {Ye, Shuo and Yu, Shujian and Wang, Yu and You, Xinge},
  title   = {{R2-trans}: Fine-grained visual categorization with redundancy reduction},
  journal = {Image and Vision Computing},
  volume  = {143},
  pages   = {104923},
  year    = {2024}
}

@inproceedings{you2021logme,
  author    = {You, Kaichao and Liu, Yong and Wang, Jianmin and Long, Mingsheng},
  title     = {{LogME}: Practical Assessment of Pre-trained Models for Transfer Learning},
  booktitle = {Proceedings of the 38th International Conference on Machine Learning},
  series    = {Proceedings of Machine Learning Research},
  volume    = {139},
  pages     = {12133--12143},
  publisher = {PMLR},
  year      = {2021}
}

@inproceedings{pandy2022gbc,
  author    = {P{\'a}ndy, Michal and Agostinelli, Andrea and Uijlings, Jasper and Ferrari, Vittorio and Mensink, Thomas},
  title     = {Transferability Estimation using {Bhattacharyya} Class Separability},
  booktitle = {Proceedings of the IEEE/CVF Conference on Computer Vision and Pattern Recognition (CVPR)},
  pages     = {9162--9172},
  year      = {2022}
}

@inproceedings{agostinelli2022stable,
  author    = {Agostinelli, Andrea and P{\'a}ndy, Michal and Uijlings, Jasper and Mensink, Thomas and Ferrari, Vittorio},
  title     = {How Stable Are Transferability Metrics Evaluations?},
  booktitle = {Proceedings of the European Conference on Computer Vision (ECCV)},
  pages     = {303--321},
  year      = {2022}
}

@inproceedings{koh2020concept,
  title={Concept bottleneck models},
  author={Koh, Pang Wei and Nguyen, Thao and Tang, Yew Siang and Mussmann, Stephen and Pierson, Emma and Kim, Been and Liang, Percy},
  booktitle={International conference on machine learning},
  pages={5338--5348},
  year={2020},
}

@misc{east2026sentinelbeetles,
  author = {Alyson East and Michael Belitz and Leah Cotton and Jacqueline Dominguez and Isabelle Betancourt and 
    S M Rayeed and Fangxun Liu and David Carlyn and Connor Kilrain and Jiaman Wu and Chandra Earl and Hilmar Lapp 
    and Kayla I. Perry and Charles Stewart and Matthew J. Thompson and Elizabeth G. Campolongo and Wei-Lun Chao and
    Eric R. Sokol and Sydne Record},
  title = {Beetles as Sentinel Taxa: Predicting drought conditions from {NEON} specimen imagery (Revision 5026be7)},
  year = {2026},
  url = {https://huggingface.co/datasets/imageomics/sentinel-beetles},
  doi = {10.57967/hf/8716},
  publisher = {Hugging Face}
}

@inproceedings{caron2021emerging,
  author    = {Caron, Mathilde and Touvron, Hugo and Misra, Ishan and J{\'e}gou, Herv{\'e} and Mairal, Julien and Bojanowski, Piotr and Joulin, Armand},
  title     = {Emerging Properties in Self-Supervised Vision Transformers},
  booktitle = {Proceedings of the IEEE/CVF International Conference on Computer Vision (ICCV)},
  pages     = {9650--9660},
  year      = {2021}
}

@inproceedings{campello2013density,
  author    = {Campello, Ricardo J. G. B. and Moulavi, Davoud and Sander, Joerg},
  title     = {Density-Based Clustering Based on Hierarchical Density Estimates},
  booktitle = {Advances in Knowledge Discovery and Data Mining (PAKDD 2013)},
  series    = {Lecture Notes in Computer Science},
  publisher = {Springer},
  pages     = {160--172},
  year      = {2013}
}

@techreport{wah2011caltech,
  title={{The Caltech-UCSD Birds-200-2011 Dataset}},
  author={Wah, Catherine and Branson, Steve and
          Welinder, Peter and Perona, Pietro and
          Belongie, Serge},
  institution={California Institute of Technology},
  number={CNS-TR-2011-001},
  year={2011}
}
